\documentclass[11pt,a4paper]{article}
\usepackage[hyperref]{acl2021}
\usepackage{times}
\usepackage{latexsym}

\usepackage{graphicx}
\usepackage{caption}
\usepackage{subcaption}
\usepackage{amsmath}
\usepackage{multirow}
\newcommand{\bos}{\textless s\textgreater}
\newcommand{\eos}{\textless /s\textgreater}

\usepackage{tikz}
\usepackage{etoolbox}
\newcommand*{\lowmark}[1]{\lower.85em \hbox{\tikz\draw (0pt, 0pt)%
    circle (.5em) node {\makebox[0.5em][c]{\tiny #1}};}}
\robustify{\lowmark}
\usepackage{soul}
\newcommand{\uline}[1]{\ul{#1}}

\usepackage[utf8]{inputenc}
\UseRawInputEncoding

\usepackage{microtype}

\aclfinalcopy 
\def\aclpaperid{1821} 

\title{G-Transformer for Document-level Machine Translation}

\author{
    Guangsheng Bao\textsuperscript{\rm 1,2},
    Yue Zhang\thanks{* Corresponding author.} \textsuperscript{\rm ,1,2},
    Zhiyang Teng\textsuperscript{\rm 1,2},
    Boxing Chen\textsuperscript{\rm 3} and
    Weihua Luo\textsuperscript{\rm 3}
    \\
    \textsuperscript{1} School of Engineering, Westlake University \\
    \textsuperscript{2} Institute of Advanced Technology, Westlake Institute for Advanced Study \\
    \textsuperscript{3} DAMO Academy, Alibaba Group Inc. \\
    \texttt{\{baoguangsheng, zhangyue, tengzhiyang\}@westlake.edu.cn} \\
    \texttt{\{boxing.cbx, weihua.luowh\}@alibaba-inc.com}
}

\date{}

\begin{document}
\maketitle
\begin{abstract}
Document-level MT models are still far from satisfactory. Existing work extend  translation unit from single sentence to multiple sentences. However, study shows that when we further enlarge the translation unit to a whole document, supervised training of Transformer can fail. In this paper, we find such failure is not caused by overfitting, but by sticking around local minima during training. Our analysis shows that the increased complexity of target-to-source attention is a reason for the failure. As a solution, we propose G-Transformer, introducing locality assumption as an inductive bias into Transformer, reducing the hypothesis space of the attention from target to source. Experiments show that G-Transformer converges faster and more stably than Transformer, achieving new state-of-the-art BLEU scores for both non-pretraining and pre-training settings on three benchmark datasets.

\end{abstract}

\section{Introduction}
Document-level machine translation (MT) has received increasing research attention \cite{gong2011cache,hardmeier2013docent,garcia2015document,werlen2018document,Maruf2019-SAN,Liu2020-mBART}. It is a more practically useful task compared to sentence-level MT because typical inputs in MT applications are text documents rather than individual sentences. A salient difference between document-level MT and sentence-level MT is that for the former, much larger inter-sentential context should be considered when translating each sentence, which include discourse structures such as anaphora, lexical cohesion, etc. Studies show that human translators consider such contexts when conducting document translation \cite{hardmeier2014discourse,laubli2018has}. Despite that neural models achieve competitive performances on sentence-level MT, the performance of document-level MT is still far from satisfactory.

Existing methods can be mainly classified into two categories. The first category translates a document sentence by sentence using a sequence-to-sequence neural model \cite{zhang-2018-improving,Miculicich2018-HAN,Maruf2019-SAN,Zheng2020-towards}. Document-level context is integrated into sentence-translation by introducing additional context encoder. The structure of such a model is shown in Figure 1(a). These methods suffer from two limitations. First, the context needs to be encoded separately for translating each sentence, which adds to the runtime complexity. Second, more importantly, information exchange cannot be made between the current sentence and its document context in the same encoding module.

\begin{figure}[t]
    \centering
    \begin{subfigure}[b]{1\linewidth}
        \centering
        \includegraphics[width=1\linewidth]{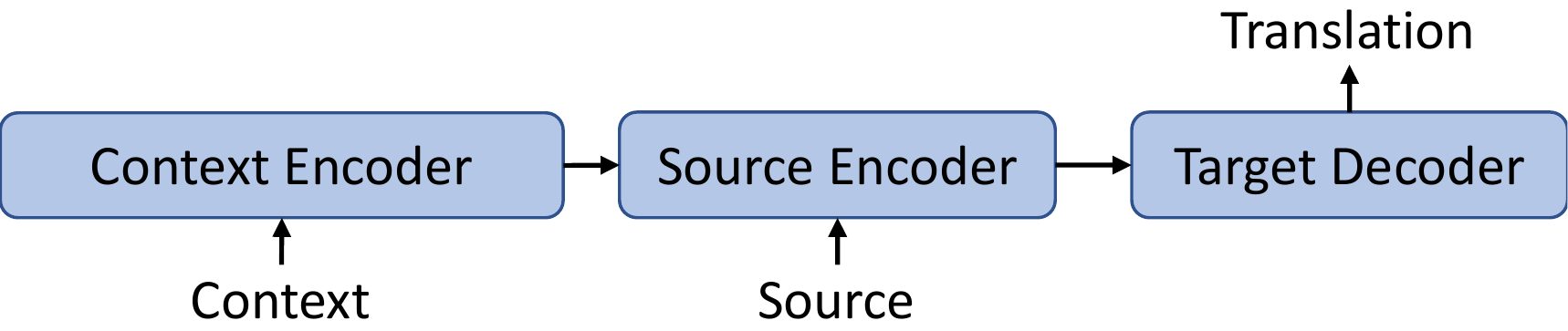}
        \caption{Sentence-by-sentence Translation}
        \label{fig:intro-sentbysent}
    \end{subfigure}
    \vfill\vspace{8pt}
    \begin{subfigure}[b]{1\linewidth}
        \centering
        \includegraphics[width=1\linewidth]{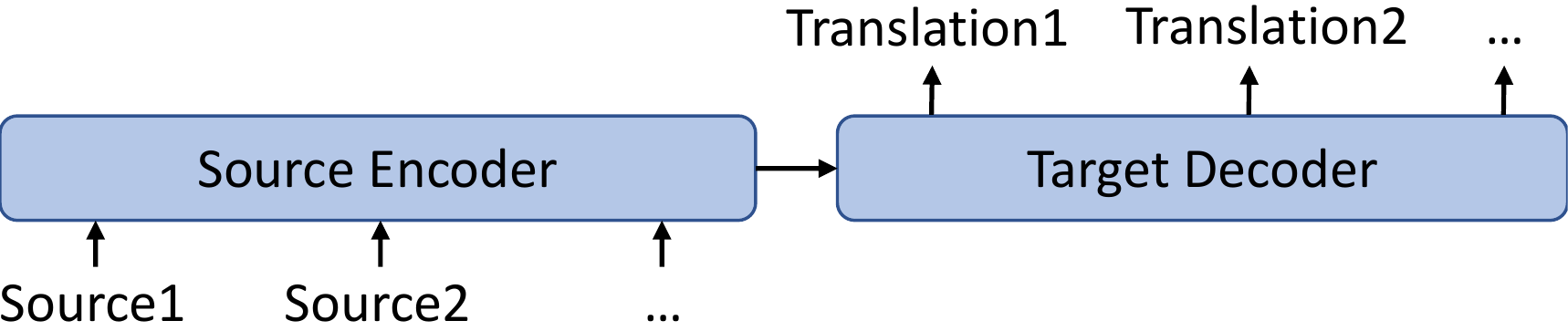}
        \caption{Multi-sentence Translation}
        \label{fig:intro-multisent}
    \end{subfigure}
    \vfill\vspace{8pt}
    \begin{subfigure}[b]{1\linewidth}
        \centering
        \includegraphics[width=1\linewidth]{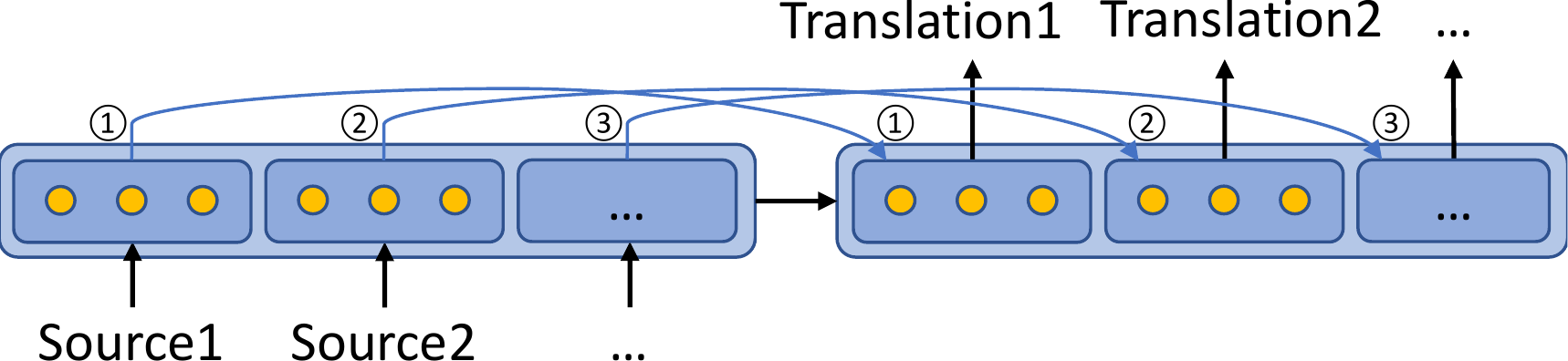}
        \caption{G-Transformer (Doc-by-doc Translation)}
        \label{fig:intro-gtrans}
    \end{subfigure}
    \caption{Overview of model structures for document-level machine translation.}
    \label{fig:model-intro}
\end{figure}

The second category extends the translation unit from a single sentence to multiple sentences \cite{tiedemann-2017-extctx,agrawal2018-contextual,zhang2020long} and the whole document \cite{junczys-2019-msdoc,Liu2020-mBART}. Recently, it has been shown that when the translation unit increases from one sentence to four sentences, the performance improves \cite{zhang2020long,scherrer-2019-anaconcat}. However, when the whole document is encoded as a single unit for sequence to sequence translation, direct supervised training has been shown to fail \cite{Liu2020-mBART}. As a solution, either large-scale pre-training \cite{Liu2020-mBART} or data augmentation \cite{junczys-2019-msdoc} has been used as a solution, leading to improved performance. These methods are shown in Figure 1(b). One limitation of such methods is that they require much more training time due to the necessity of data augmentation.

Intuitively, encoding the whole input document as a single unit allows the best integration of context information when translating the current sentence. However, little work has been done investigating the underlying reason why it is difficult to train such a document-level NMT model. One remote clue is that as the input sequence grows larger, the input becomes more sparse \cite{pouget2014overcoming,koehn2017six}. To gain more understanding, we make dedicated experiments on the influence of input length, data scale and model size for Transformer (Section \ref{sec:trans_analysis}), finding that a Transformer model can fail to converge when training with long sequences, small datasets, or big model size.
We further find that for the failed cases, the model gets stuck at local minima during training. In such situation, the attention weights from the decoder to the encoder are flat, with large entropy values. This can be because that larger input sequences increase the challenge for focusing on a local span to translate when generating each target word. In other words, the hypothesis space for target-to-source attention is increased.

Given the above observations, we investigate a novel extension of Transformer, by restricting self-attention and target-to-source attention to a local context using a guidance mechanism. As shown in Figure 1(c), while we still encode the input document as a single unit, group tags \textcircled{\scriptsize 1} \textcircled{\scriptsize 2} \textcircled{\scriptsize 3}  are assigned to sentences to differentiate their positions. Target-to-source attention is guided by matching the tag of target sentence to the tags of source sentences when translating each sentence, so that the hypothesis space of attention is reduced. Intuitively, the group tags serve as a constraint on attention, which is useful for differentiating the current sentence and its context sentences. Our model, named G-Transformer, can be thus viewed as a combination of the method in Figure 1(a) and Figure 1(b), which fully separate and fully integrates a sentence being translated with its document level context, respectively.

We evaluate our model on three commonly used document-level MT datasets for English-German translation, covering domains of TED talks, News, and Europarl from small to large. Experiments show that G-Transformer converges faster and more stably than Transformer on different settings, obtaining the state-of-the-art results under both non-pretraining and pre-training settings. 
To our knowledge, we are the first to realize a truly document-by-document translation model. We release our code and model at https://github.com/baoguangsheng/g-transformer.

\begin{table*}[t]
    \centering\small
    \begin{tabular}{ll|rrr|rr}
        \hline
        \multirow{2}{*}{\bf{Language}} & \multirow{2}{*}{\bf{Dataset}} & \bf{\#Sentences} & \bf{\#Documents} & \bf{\#Instances} & \bf{Avg \#Sents/Inst} & \bf{Avg \#Tokens/Inst} \\
         & & \bf{train/dev/test} & \bf{train/dev/test} & \bf{train/dev/test} & \bf{train/dev/test} & \bf{train/dev/test} \\
        \hline
        \multirow{3}{*}{En-De} & TED & 0.21M/9K/2.3K  &  1.7K/92/22  &  11K/483/123 & 18.3/18.5/18.3 & 436/428/429 \\
        & News & 0.24M/2K/3K  &  6K/80/154  &  18.5K/172/263 & 12.8/12.6/11.3 & 380/355/321 \\
        & Europarl & 1.67M/3.6K/5.1K  & 118K/239/359  &  162K/346/498 & 10.3/10.4/10.3 & 320/326/323 \\
        \hline
    \end{tabular}
    \caption{En-De datasets for evaluation.}
    \label{tab:datasets}
\end{table*}

\section{Experimental Settings}
\label{sec:exp_settings}

We evaluate Transformer and G-Transformer on the widely adopted \emph{benchmark datasets} \cite{Maruf2019-SAN}, including three domains for English-German (En-De) translation.

\textbf{TED.}
The corpus is transcriptions of TED talks from IWSLT 2017. Each talk is used as a document, aligned at the sentence level. {\it tst2016-2017} is used for testing, and the rest for development.

\textbf{News.}
This corpus uses News Commentary v11 for training, which is document-delimited and sentence-aligned. {\it newstest2015} is used for development, and {\it newstest2016} for testing.

\textbf{Europarl.}
The corpus is extracted from Europarl v7, where sentences are segmented and aligned using additional information. The train, dev and test sets are randomly split from the corpus.

The detailed statistics of these corpora are shown in Table \ref{tab:datasets}. We pre-process the documents by splitting them into instances with up-to $512$ tokens, taking a sentence as one instance if its length exceeds $512$ tokens.
We tokenize and truecase the sentences with MOSES \cite{koehn2007moses} tools, applying BPE \cite{sennrich2016neural} with $30000$ merging operations.

We consider three standard \emph{model configurations}.

\textbf{Base Model.}
Following the standard Transformer base model \cite{vaswani2017attention}, we use 6 layers, 8 heads, 512 dimension outputs, and 2048 dimension hidden vectors.

\textbf{Big Model.}
We follow the standard Transformer big model \cite{vaswani2017attention}, using 6 layers, 16 heads, 1024 dimension outputs, and 4096 dimension hidden vectors.

\textbf{Large Model.}
We use the same settings of BART large model \cite{Lewis2020-BART}, which involves 12 layers, 16 heads, 1024 dimension outputs, and 4096 dimension hidden vectors.

We use s-BLEU and d-BLEU \cite{Liu2020-mBART} as the \emph{metrics}.
The detailed descriptions are in Appendix \ref{apdx:metrics}.

\section{Transformer and Long Inputs}
\label{sec:trans_analysis}
We empirically study Transformer (see Appendix \ref{apdx:transformer}) on the datasets. 
We run each experiment five times using different random seeds, reporting the average score for comparison.

\begin{figure}[t]
    \centering
    \begin{subfigure}[b]{0.49\linewidth}
        \centering
        \includegraphics[width=1\linewidth]{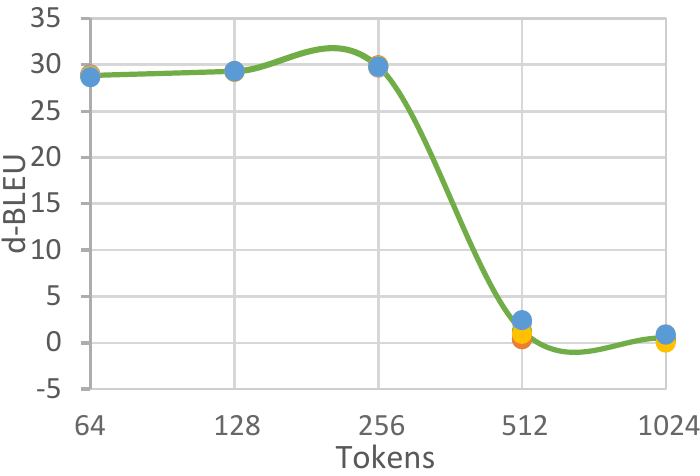}
        \caption{Input Length (Base model with filtered data.)}
        \label{fig:trans-inputsize}
    \end{subfigure}
    \hfill
    \begin{subfigure}[b]{0.49\linewidth}
        \centering
        \includegraphics[width=1\linewidth]{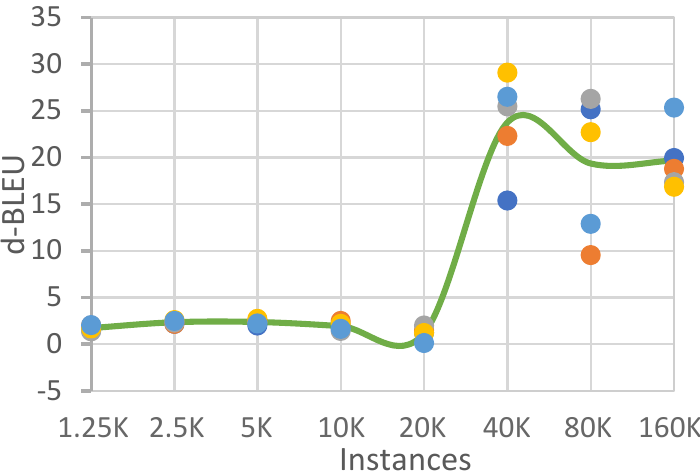}
        \caption{Data Scale (Base model with 512 tokens input.)}
        \label{fig:trans-datasize}
    \end{subfigure}
    \caption{Transformer on various input length and data scale.}
    \label{fig:trans-problems}
\end{figure}

\subsection{Failure Reproduction}

\textbf{Input Length.} 
We use the Base model and fixed dataset for this comparison. We split both the training and testing documents from Europarl dataset into instances with input length of 64, 128, 256, 512, and 1024 tokens, respectively. For fair comparison, we remove the training documents with a length of less than 768 tokens, which may favour small input length.
The results are shown in Figure \ref{fig:trans-inputsize}. When the input length increases from 256 tokens to 512 tokens, the BLEU score drops dramatically from 30.5 to 2.3, indicating failed training with 512 and 1024 tokens. It demonstrates the difficulty when dealing with long inputs of Transformer.

\textbf{Data Scale.} 
We use the Base model and a fixed input length of 512 tokens. For each setting, we randomly sample a training dataset of the expected size from the full dataset of Europarl. The results are shown in Figure \ref{fig:trans-datasize}. The performance increases sharply when the data scale increases from 20K to 40K. When data scale is equal or less than 20K, the BLEU scores are under 3, which is unreasonably low, indicating that with a fixed model size and input length, the smaller dataset can also cause the failure of the training process. For data scale more than 40K, the BLEU scores show a wide dynamic range, suggesting that the training process is unstable.

\textbf{Model Size.} 
We test Transformer with different model sizes, using the full dataset of Europarl and a fixed input length of 512 tokens.
Transformer-Base can be trained successfully, giving a reasonable BLEU score.  However, the training of the Big and Large models failed, resulting in very low BLEU scores under 3. It demonstrates that the increased model size can also cause the failure with a fixed input length and data scale.

The results confirm the intuition that the performance will drop with longer inputs, smaller datasets, or bigger models.
However, the BLEU scores show a strong discontinuity with the change of input length, data scale, or model size, falling into two discrete clusters. One is successfully trained cases with d-BLEU scores above 10, and the other is failed cases with d-BLEU scores under 3.

\begin{figure}
    \centering
    \begin{subfigure}[b]{0.485\linewidth}
        \centering
        \includegraphics[width=1\linewidth]{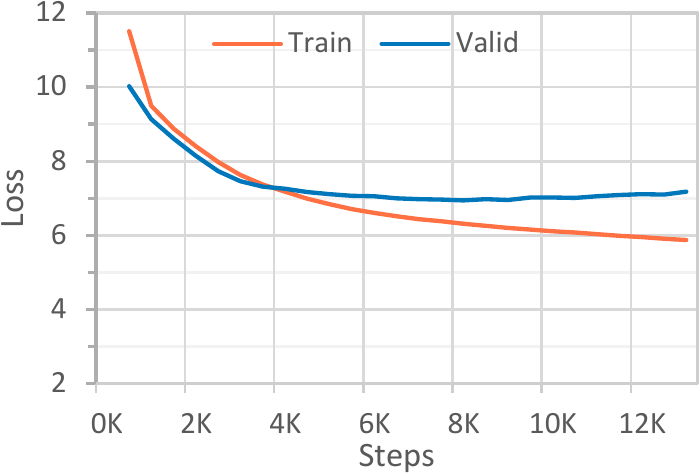}
        \caption{Failed Model}
        \label{fig:trans-loss-failed}
    \end{subfigure}
    \begin{subfigure}[b]{0.495\linewidth}
        \centering
        \includegraphics[width=1\linewidth]{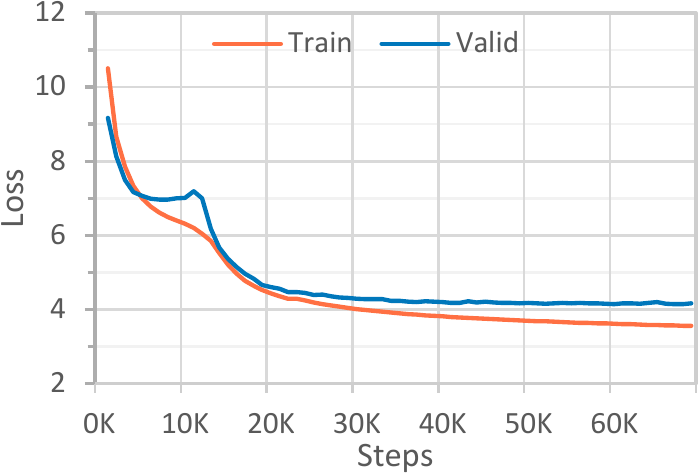}
        \caption{Successful Model}
        \label{fig:trans-loss-success}
    \end{subfigure}
    \caption{Loss curve of the models and the local minima.}
    \label{fig:trans-lossanalysis}
\end{figure}

\subsection{Failure Analysis}

\textbf{Training Convergence.}
Looking into the failed models, we find that they have a similar pattern on loss curves. As an example of the model trained on 20K instances shown in Figure \ref{fig:trans-loss-failed}, although the training loss continually decreases during training process, the validation loss sticks at the level of 7, reaching a minimum value at around 9K training steps.
In comparison, the successfully trained models share another pattern. Taking the model trained on 40K instances as an example, the loss curves demonstrate two stages, which is shown in Figure \ref{fig:trans-loss-success}. In the first stage, the validation loss similar to the failed cases has a converging trend to the level of 7. In the second stage, after 13K training steps, the validation loss falls suddenly, indicating that the model may escape successfully from local minima.
From the two stages of the learning curve, we conclude that the real problem, contradicting our first intuition, is not about overfitting, but about local minima.

\begin{figure}
    \centering
    \begin{subfigure}[b]{0.49\linewidth}
        \centering
        \includegraphics[width=1\linewidth]{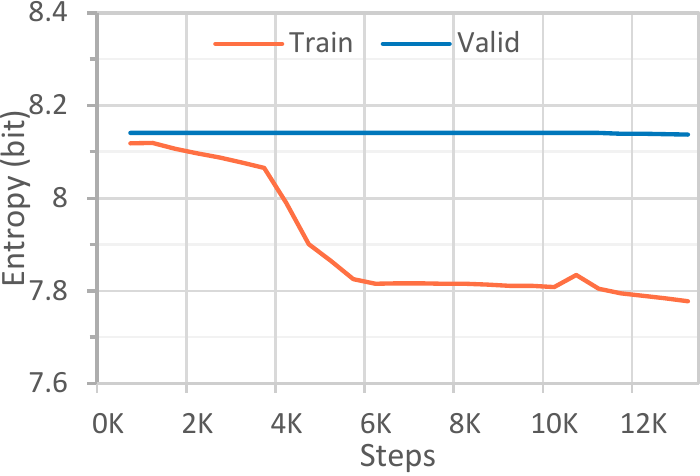}
        \caption{Failed Model}
        \label{fig:trans-crossattn-failed}
    \end{subfigure}
    \hfill
    \begin{subfigure}[b]{0.49\linewidth}
        \centering
        \includegraphics[width=1\linewidth]{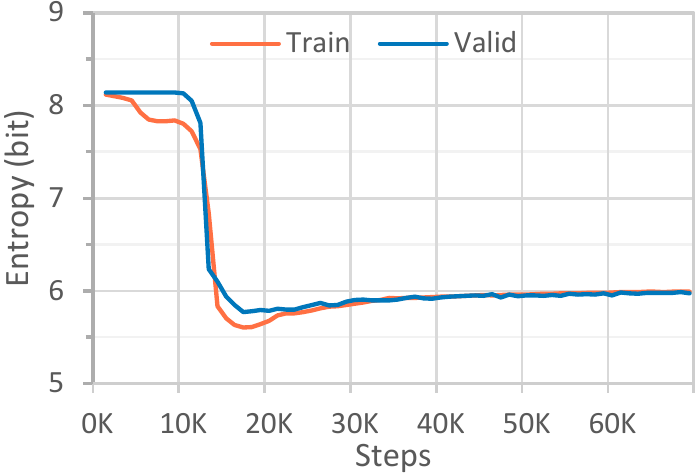}
        \caption{Successful Model}
        \label{fig:trans-crossattn-success}
    \end{subfigure}
    \caption{Cross-attention distribution of Transformer shows that the failed model sticks at the local minima.}
    \label{fig:trans-crossattn}
\end{figure}

\begin{figure}
    \centering
    \begin{subfigure}[b]{0.49\linewidth}
        \centering
        \includegraphics[width=1\linewidth]{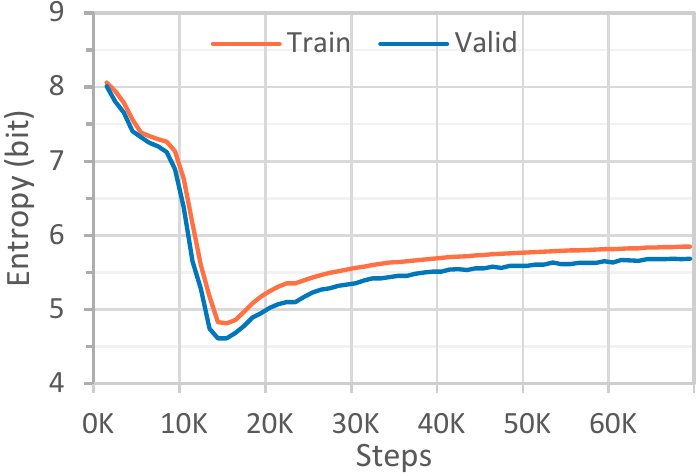}
        \caption{Encoder}
        \label{fig:trans-selfattn-encoder}
    \end{subfigure}
    \hfill
    \begin{subfigure}[b]{0.49\linewidth}
        \centering
        \includegraphics[width=1\linewidth]{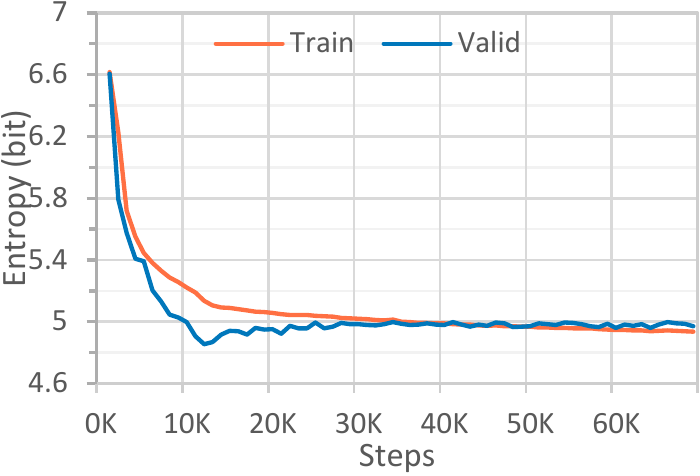}
        \caption{Decoder}
        \label{fig:trans-selfattn-decoder}
    \end{subfigure}
    \caption{For the successful model, the attention distribution shrinks to narrow range (low entropy) and then expands to wider range (high entropy).}
    \label{fig:trans-selfattn}
\end{figure}

\textbf{Attention Distribution.}
We further look into the attention distribution of the failed models, observing that the attentions from target to source are widely spread over all tokens. As Figure \ref{fig:trans-crossattn-failed} shows, the distribution entropy is high for about 8.14 bits on validation. In contrast, as shown in Figure \ref{fig:trans-crossattn-success}, the successfully trained model has a much lower attention entropy of about 6.0 bits on validation. Furthermore, we can see that before 13K training steps, the entropy sticks at a plateau, confirming with the observation of the local minima in Figure \ref{fig:trans-loss-success}. It indicates that the early stage of the training process for Transformer is difficult.

Figure \ref{fig:trans-selfattn} shows the self-attention distributions of the successfully trained models. The attention entropy of both the encoder and the decoder drops fast at the beginning, leading to a shrinkage of the attention range. But then the attention entropy gradually increases, indicating an expansion of the attention range. Such back-and-forth oscillation of the attention range may also result in unstable training and slow down the training process.

\subsection{Conclusion}
The above experiments show that training failure on Transformer can be caused by local minima. Additionally, the oscillation of attention range may make it worse.
During training process, the attention module needs to identify relevant tokens from whole sequence to attend to. Assuming that the sequence length is $N$, the complexity of the attention distribution increases when $N$ grows from sentence-level to document-level.

We propose to use locality properties \cite{rizzi2013locality,hardmeier2014discourse,jawahar2019does} of both the language itself and the translation task as a constraint in Transformer, regulating the hypothesis space of the self-attention and target-to-source attention, using a simple group tag method.

\begin{figure}[t]
    \centering\scriptsize
    \setlength{\fboxrule}{0.2pt}
    \setlength{\fboxsep}{0.2cm}
    \fbox{\parbox{0.95\linewidth}{
        \setlength{\baselineskip}{12pt}
        \textbf{Source: }
        \uline{{\bos} the Commission shares ... of the European Union institutional framework . {\eos}}\lowmark{1} \uline{{\bos} Commission participation is expressly provided for ... of all its preparatory bodies . {\eos}}\lowmark{2} \uline{{\bos} only in exceptional circumstances ... be excluded from these meetings . {\eos}}\lowmark{3} ...

        \vskip 0.2cm
        \textbf{Target: }
        \uline{{\bos} die Kommission teilt die Ansicht ... des institutionellen Rahmens der Europäischen Union ist . {\eos}}\lowmark{1} 
        \uline{{\bos} die Geschäftsordnung des Rates ... der Kommission damit ausdrücklich vor . {\eos}}\lowmark{2} 
        \uline{{\bos} die Kommission kann nur ... wobei fallweise zu entscheiden ist . {\eos}}\lowmark{3} ...
        \par
    }}
    \caption{Example of English-German translation with group alignments.}
    \label{fig:caseofgtrans}
\end{figure}

\section{G-Transformer}
An example of G-Transformer is shown in Figure \ref{fig:caseofgtrans}, where the input document contains more than 3 sentences. As can be seen from the figure, G-Transformer extends Transformer by augmenting the input and output with group tags \cite{bao-2021}. In particular, each token is assigned a group tag, indicating its sentential index. While source group tags can be assigned deterministically, target tags are assigned dynamically according to whether a generated sentence is complete. Starting from 1, target words copy group tags from its predecessor unless the previous token is {\eos}, in which case the tag increases by 1. The tags serve as a locality constraint, encouraging target-to-source attention to concentrate on the current source sentence being translated. 

Formally, for a source document $X$ and a target document $Y$, the probability model of Transformer can be written as
\begin{equation}
\small
  \hat{Y} = \arg \max_{Y} P(Y|X), 
\label{eq:def:trans}
\end{equation}
and G-Transformer extends it by having
\begin{equation}
\small
  \hat{Y} = \arg_Y \max_{Y,G_Y} P(Y,G_Y|X, G_X), 
\label{eq:def:gtrans}
\end{equation}
where $G_X$ and $G_Y$ denotes the two sequences of group tags
\begin{equation}
\small
  \begin{split}
  G_X &= \{g_i= k \text{ if } w_i \in sent^X_k \text{ else } 0\}|_{i=1}^{|X|}, \\
  G_Y &= \{g_j= k \text{ if } w_j \in sent^Y_k \text{ else } 0\}|_{j=1}^{|Y|}, \\
  \end{split}
\label{eq:gxgy}
\end{equation}
where $sent_k$ represents the $k$-th sentence of $X$ or $Y$. For the example shown in Figure \ref{fig:caseofgtrans}, $G_X=\{1,...,1,2,...,2,3,...,3,4,...\}$ and $G_Y=\{1,...,1,2,...,2,3,...,3,4,...\}$.

Group tags influence the auto-regressive translation process by interfering with the attention mechanism, which we show in the next section. 
In G-Transformer, we use the group-tag sequence $G_X$ and $G_Y$ for representing the alignment between $X$ and $Y$, and for generating the {\it localized} contextual representation of $X$ and $Y$.

\begin{table*}[t]
    \centering\small
    \begin{tabular}{l|cc|cc|cc}
        \hline
        \bf{Method} & \multicolumn{2}{c|}{\bf{TED}} & \multicolumn{2}{c|}{\bf{News}} & \multicolumn{2}{c}{\bf{Europarl}} \\
         & s-BLEU & d-BLEU & s-BLEU & d-BLEU & s-BLEU & d-BLEU \\
        \hline
        \textsc{SentNmt}~\cite{vaswani2017attention} & 23.10 & - & 22.40 & - & 29.40 & - \\
        HAN~\cite{Miculicich2018-HAN} & 24.58 & - & 25.03 & - & 28.60 & - \\
        SAN~\cite{Maruf2019-SAN} & 24.42 & - & 24.84 & - & 29.75 & - \\
        Hybrid Context~\cite{Zheng2020-towards} & 25.10 & - & 24.91 & - & 30.40 & - \\
        Flat-Transformer~\cite{Ma2020-flat} & 24.87 & - & 23.55 & - & 30.09 & - \\
        \hline
        Transformer on sent (baseline) & 24.82 & - & 25.19 & - & 31.37 & -  \\
        Transformer on doc (baseline) & - & 0.76 & - & 0.60 & - & 33.10  \\
        G-Transformer random initialized (ours) & 23.53 & 25.84* & 23.55 & 25.23* & 32.18* & 33.87*  \\
        G-Transformer fine-tuned on sent Transformer (ours) & \bf{25.12} & \bf{27.17}* & \bf{25.52} & \bf{27.11}* & \bf{32.39}* & \bf{34.08}*  \\
        \hline
        \multicolumn{7}{c}{Fine-tuning on Pre-trained Model} \\
        \hline
        Flat-Transformer+BERT~\cite{Ma2020-flat} & 26.61 & - & 24.52 & - & 31.99 & - \\
        G-Transformer+BERT (ours) & \bf{26.81} & - & \bf{26.14} & - & \bf{32.46} & - \\
        \hline
        Transformer on sent fine-tuned on BART (baseline) & 27.78 & - & 29.90 & - & 31.87 & -  \\
        Transformer on doc fine-tuned on BART (baseline) & - & 28.29 & - & 30.49 & - & 34.00  \\
        G-Transformer fine-tuned on BART (ours) & \bf{28.06} & \bf{30.03}* & \bf{30.34}* & \bf{31.71}* & \bf{32.74}* & \bf{34.31}*  \\
        \hline
    \end{tabular}
    \caption{Case-sensitive BLEU scores on En-De translation. ``*'' indicates statistically significant at $p<0.01$ compared to the Transformer baselines.}
    \label{tab:compare-prevdoc}
\end{table*}

\subsection{Group Attention}
An attention module can be seen as a function mapping a query and a set of key-value pairs to an output \cite{vaswani2017attention}. The query, key, value, and output are all vectors. The output is computed by summing the values with corresponding attention weights, which are calculated by matching the query and the keys.
Formally, given a set of queries, keys, and values, we pack them into matrix $Q$, $K$, and $V$, respectively. We compute the matrix outputs
\begin{equation}
\small
    \text{Attention}(Q, K, V) = \text{softmax} \left( \frac{QK^T}{\sqrt{d_k}} \right) V,
\label{eq:singlehead}
\end{equation}
where $d_k$ is the dimensions of the key vector.

Attention allows a model to focus on different positions. Further, multi-head attention (MHA) allows a model to gather information from different representation subspaces
\begin{equation}
\small
\begin{split}
    \text{MHA}(Q, K, V) &= \text{Concat}(head_1,...,head_h)W^O, \\
    head_i &= \text{Attention}(Q W_i^Q, K W_i^K, V W_i^V), \\
\end{split}
\label{eq:multihead}
\end{equation}
where the projections of $W^O$, $W_i^Q$, $W_i^K$, and $W_i^V$ are parameter matrices.

We update Eq \ref{eq:singlehead} using group-tags, naming it group attention (GroupAttn). In addition to inputs $Q$, $K$, and $V$, two sequences of group-tag inputs are involved, where $G_Q$ corresponds to $Q$ and $G_K$ corresponds to $K$. We have
\begin{equation}
\small
\begin{split}
    args &= (Q, K, V, G_Q, G_K), \\
    \text{GroupAttn}(args) &= \text{softmax} \left( \frac{QK^T}{\sqrt{d_k}} + M(G_Q, G_K) \right) V, \\
\end{split}
\end{equation}
where function $M(\cdot)$ works as an attention mask, excluding all tokens outside the sentence. Specifically, $M(\cdot)$ gives a big negative number $\gamma$ to make \emph{softmax} close to $0$ for the tokens with a different group tag compared to current token
\begin{equation}
\small
    M(G_Q, G_K) = \text{min}(1, \text{abs}(G_Q I_K^T - I_Q G_K^T)) * \gamma,
\end{equation}
where $I_K$ and $I_Q$ are constant vectors with value $1$ on all dimensions, that $I_K$ has dimensions equal to the length of $G_K$ and $I_Q$ has dimensions equal to the length of $G_Q$. The constant value $\gamma$ can typically be $-1e8$.

Similar to Eq \ref{eq:multihead}, we use group multi-head attention
\begin{equation}
\small
\begin{split}
    args &= (Q, K, V, G_Q, G_K), \\
    \text{GroupMHA}(args) &= \text{Concat}(head_1,...,head_h)W^O, \\
\end{split}
\label{eq:groupmha}
\end{equation}
where
\begin{equation}
\small
    head_i = \text{GroupAttn}(Q W_i^Q, K W_i^K, V W_i^V, G_Q, G_K), \\    
\end{equation}
and the projections of $W^O$, $W_i^Q$, $W_i^K$, and $W_i^V$ are parameter matrices.

\textbf{Encoder.}
For each layer a group multi-head attention module is used for self-attention, assigning the same group-tag sequence for the key and the value that $G_Q=G_K=G_X$.

\textbf{Decoder.}
We use one group multi-head attention module for self-attention and another group multi-head attention module for cross-attention. Similar to the encoder, we assign the same group-tag sequence to the key and value of the self-attention, that $G_Q=G_K=G_Y$, but use different group-tag sequences for cross-attention that $G_Q=G_Y$ and $G_K=G_X$.

\textbf{Complexity.}
Consider a document with $M$ sentences and $N$ tokens, where each sentence contains $N/M$ tokens on average. The complexities of both the self-attention and cross-attention in Transformer are $O(N^2)$. In contrast, the complexity of group attention in G-Transformer is $O(N^2/M)$ given the fact that the attention is restricted to a local sentence.
Theoretically, since the average length $N/M$ of sentences tends to be constant, the time and memory complexities of group attention are approximately $O(N)$, making training and inference on very long inputs feasible.

\subsection{Combined Attention}
We use only group attention on lower layers for local sentence representation, and combined attention on top layers for integrating local and global context information. We use the standard multi-head attention in Eq \ref{eq:multihead} for global context, naming it global multi-head attention (GlobalMHA). Group multi-head attention in Eq \ref{eq:groupmha} and global multi-head attention are combined using a gate-sum module \cite{zhang2016gated,tu2017context}
\begin{equation}
\small
\begin{split}
    H_L &= \text{GroupMHA}(Q, K, V, G_Q, G_K), \\
    H_G &= \text{GlobalMHA}(Q, K, V), \\
    g &= \text{sigmoid}([H_L, H_G]W + b), \\    
    H &= H_L \odot g + H_G \odot (1 - g), \\
\end{split}
\label{eq:combineattn}
\end{equation}
where $W$ and $b$ are linear projection parameters, and $\odot$ denotes element-wise multiplication.

Previous study \cite{jawahar2019does} shows that the lower layers of Transformer catch more local syntactic relations, while the higher layers represent longer distance relations. Based on these findings, we use combined attention only on the top layers for integrating local and global context. By this design, on lower layers, the sentences are isolated from each other, while on top layers, the cross-sentence interactions are enabled. Our experiments show that the top 2 layers with global attention are sufficient for document-level NMT, and more layers neither help nor harm the performance. 

\subsection{Inference}
During decoding, we generate group-tag sequence $G_Y$ according to the predicted token, starting with $1$ at the first {\bos} and increasing $1$ after each {\eos}.
We use beam search and apply the maximum length constraint on each sentence. We generate the whole document from start to end in one beam search process, using a default beam size of 5.

\section{G-Transformer Results}
We compare G-Transformer with Transformer baselines and  previous document-level NMT models on both non-pretraining and pre-training settings. The detailed descriptions about these training settings are in Appendix \ref{apdx:gtrans-training}. We make statistical significance test according to \citet{collins2005clause}.

\subsection{Results on Non-pretraining Settings}
As shown in Table \ref{tab:compare-prevdoc}, the sentence-level Transformer outperforms previous document-level models on News and Europarl. Compared to this strong baseline, our randomly initialized model of G-Transformer improves the s-BLEU by 0.81 point on the large dataset Europarl. The results on the small datasets TED and News are worse, indicating overfitting with long inputs. When G-Transformer is trained by fine-tuning the sentence-level Transformer, the performance improves on the three datasets by 0.3, 0.33, and 1.02 s-BLEU points, respectively.

Different from the baseline of document-level Transformer, G-Transformer can be successfully trained on small TED and News. On Europarl, G-Transformer outperforms Transformer by 0.77 d-BLEU point, and   G-Transformer fine-tuned on sentence-level Transformer enlarges the gap to 0.98 d-BLEU point.

G-Transformer outperforms previous document-level MT models on News and Europarl with a significant margin. Compared to the best recent model Hyrbid-Context, G-Transformer improves the s-BLEU on Europarl by 1.99. These results suggest that in contrast to previous short-context models, sequence-to-sequence model taking the whole document as input is a promising direction. 

\subsection{Results on Pre-training Settings}
There is relatively little existing work about document-level MT using pre-training. Although Flat-Transformer+BERT gives a state-of-the-art scores on TED and Europarl, the score on News is worse than previous non-pretraining model HAN \cite{Miculicich2018-HAN}. G-Transformer+BERT improves the scores by margin of 0.20, 1.62, and 0.47 s-BLEU points on TED, News, and Europarl, respectively. It shows that with a better contextual representation, we can further improve document-level MT on pretraining settings.

We further build much stronger Transformer baselines by fine-tuning on mBART25 \cite{Liu2020-mBART}. 
Taking advantage of sequence-to-sequence pre-training, the sentence-level Transformer gives much better s-BLEUs of 27.78, 29.90, and 31.87, respectively.
G-Transformer fine-tuned on mBART25 improves the performance by 0.28, 0.44, and 0.87 s-BLEU, respectively. Compared to the document-level Transformer baseline, G-Transformer gives 1.74, 1.22, and 0.31 higher d-BLEU points, respectively. It demonstrates that even with well-trained sequence-to-sequence model, the locality bias can still enhance the performance.

\begin{figure}[t]
    \centering
    \setlength{\belowcaptionskip}{-6pt}    
    \begin{subfigure}[b]{0.49\linewidth}
        \centering
        \includegraphics[width=1\linewidth]{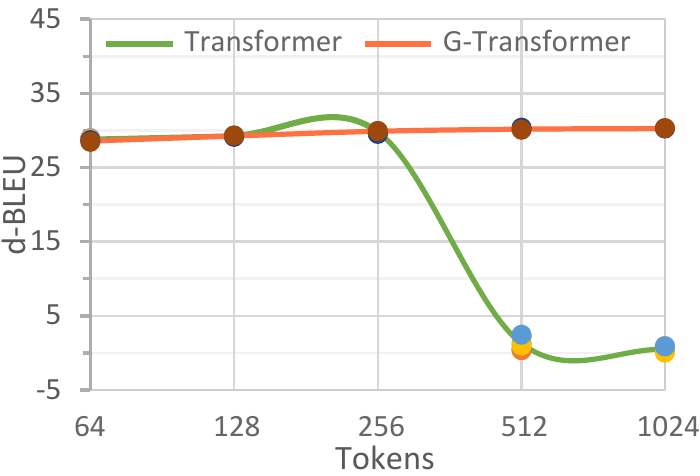}
        \caption{Input Length}
        \label{fig:gtrans-inputsize}
    \end{subfigure}
    \hfill
    \begin{subfigure}[b]{0.49\linewidth}
        \centering
        \includegraphics[width=1\linewidth]{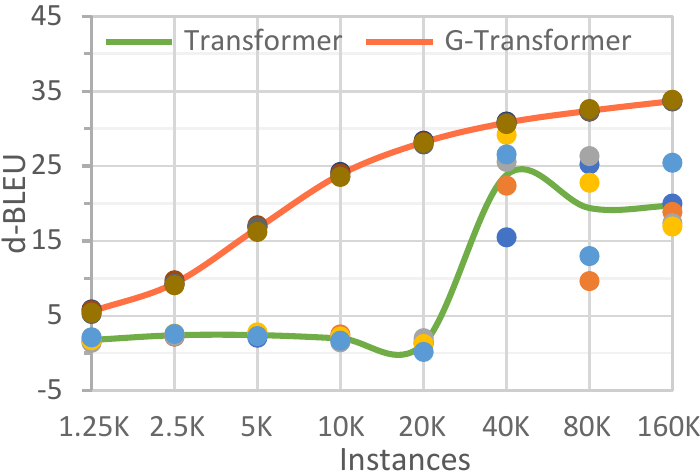}
        \caption{Data Scale}
        \label{fig:gtrans-datasize}
    \end{subfigure}
    \caption{G-Transformer compared with Transformer.}
    \label{fig:compare-transmodels}
\end{figure}

\subsection{Convergence}
We evaluate G-Transformer ad Transformer on various input length, data scale, and model size to better understand that to what extent it has solved the convergence problem of Transformer.

\begin{figure}[t]
    \centering
    \setlength{\belowcaptionskip}{-6pt}    
    \begin{subfigure}[b]{0.49\linewidth}
        \centering
        \includegraphics[width=1\linewidth]{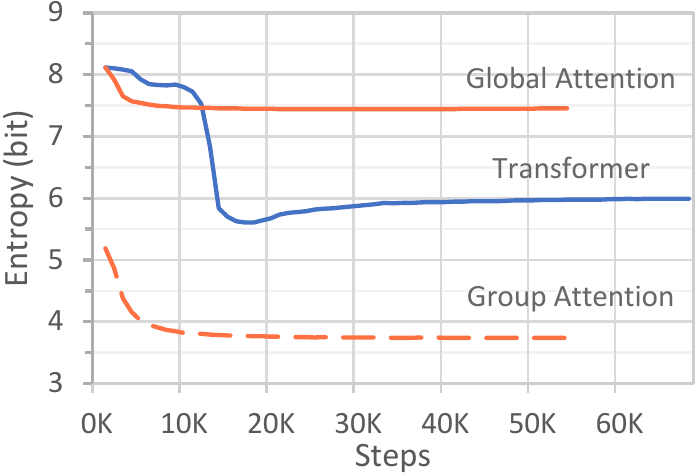}
        \caption{Cross-Attention}
        \label{fig:gtrans-crossattn-compare}
    \end{subfigure}
    \hfill
    \begin{subfigure}[b]{0.49\linewidth}
        \centering
        \includegraphics[width=1\linewidth]{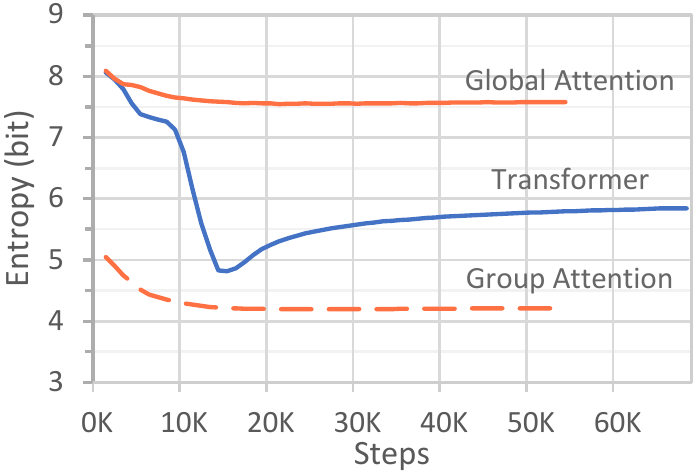}
        \caption{Encoder Self-Attention}
        \label{fig:gtrans-encoderselfattn-compare}
    \end{subfigure}
    \caption{Comparison on the development of cross-attention and encoder self-attention.}
    \label{fig:gtrans-attncompare}
\end{figure}

\textbf{Input Length.}
The results are shown in Figure \ref{fig:gtrans-inputsize}. Unlike Transformer, which fails to train on long input, G-Transformer shows stable scores for inputs containing 512 and 1024 tokens, suggesting that with the help of locality bias, a long input does not impact the performance obviously.

\textbf{Data Scale.} 
As shown in Figure \ref{fig:gtrans-datasize}, overall G-Transformer has a smooth curve of performance on the data scale from 1.25K to 160K. The variances of the scores are much lower than Transformer, indicating stable training of G-Transformer. Additionally, G-Transformer outperforms Transformer by a large margin on all the settings.

\textbf{Model Size.} 
Unlike Transformer, which fails to train on Big and Large model settings, G-Transformer shows stable scores on different model sizes.
As shown in Appendix \ref{apdx:gtrans-modelsize}, although performance on small datasets TED and News drops largely for Big and Large model, the performance on large dataset Europarl only decreases by 0.10 d-BLEU points for the Big model and 0.66 for the Large model.

\textbf{Loss.} 
Looking into the training process of the above experiments, we see that both the training and validation losses of G-Transformer converge much faster than Transformer, using almost half time to reach the same level of loss. Furthermore, the validation loss of G-Transformer converges to much lower values. These observations demonstrate that G-Transformer converges faster and better.

\textbf{Attention Distribution.}
Benefiting from the separate group attention and global attention, G-Transformer avoids the oscillation of attention range, which happens to Transformer. As shown in Figure \ref{fig:gtrans-crossattn-compare}, Transformer sticks at the plateau area for about 13K training steps, but G-Transformer shows a quick and monotonic convergence, reaching the stable level using about 1/4 of the time that Transformer takes. Through Figure \ref{fig:gtrans-encoderselfattn-compare}, we can find that G-Transformer also has a smooth and stable curve for the convergence of self-attention distribution. These observations imply that the potential conflict of local sentence and document context can be mitigated by G-Transformer.

\begin{table}[t]
    \centering\small
    \begin{tabular}{@{}l|ccc|c@{}}
        \hline
        \bf{Method} & \bf{TED} & \bf{News} & \bf{Europarl} & \bf{Drop}\\
        \hline
        G-Transformer (fnt.) & 25.12 & 25.52 & 32.39 & - \\
         - target-side context & 25.05 & 25.41 & 32.16 & -0.14 \\
         - source-side context & 24.56 & 24.58 & 31.39 & -0.70 \\
         \hline
    \end{tabular}
    \caption{Impact of source-side and target-side context reporting in s-BLEU. Here, fnt. denotes the model fine-tuned on sentence-level Transformer.}
    \label{tab:context}
\end{table}

\begin{table}[t]
    \centering\small
    \begin{tabular}{@{}l|ccc@{}}
        \hline
        \bf{Method} & \bf{deixis} & \bf{el.infl.} & \bf{el.VP} \\
        \hline
        CADec \cite{voita2019good} & 81.6 & 72.2 & 80.0  \\
        LSTM-Tran \cite{zhang2020long} & \bf{91.0} & 82.2 & 78.2 \\
        \hline
        sent \cite{voita2019good} & 50.0 & 53.0 & 28.4 \\
        concat \cite{voita2019good} & 83.5 & 76.2 & 76.6 \\
        G-Transformer & 89.9 & \bf{84.8} & \bf{82.4} \\
        \hline
    \end{tabular}
    \caption{Impact on discourse by the source-side context, in accuracy of correctly identifying the discourse phenomena. Here, el. means ellipsis. LSTM-Tran denotes LSTM-Transformer.}
    \label{tab:discourse}
\end{table}

\begin{table}[t]
    \centering\small
    \begin{tabular}{@{}l|cc@{\hskip 10pt}c@{ }|c@{}}
        \hline
        \bf{Method} & \bf{TED} & \bf{News} & \bf{Europarl} & \bf{Drop}\\
        \hline
        G-Transformer (rnd.) & 25.84 & 25.23 & 33.87 & - \\
         - word-dropout & 25.49 & 24.65 & 33.70 & -0.37 \\        
         - language locality & 22.47 & 22.41 & 33.63 & -1.78 \\
         - translation locality & 0.76 & 0.60 & 33.10 & -14.68 \\
         \hline
    \end{tabular}
    \caption{Contribution of locality bias and word-dropout reporting in d-BLEU. Here, rnd. denotes the model trained using randomly initialized parameters.}
    \label{tab:ablation}
\end{table}

\begin{table}[t]
    \centering\small
    \begin{tabular}{@{}l|ccc@{ }|c@{}}
        \hline
        \bf{Method} & \bf{TED} & \bf{News} & \bf{Europarl} & \bf{Drop}\\
        \hline
        \multicolumn{5}{c}{G-Transformer (rnd.)} \\
        \hline
        Combined attention & 25.84 & 25.23 & 33.87 & - \\
        Only group attention & 25.62 & 25.14 & 33.12 & -0.35 \\        
        Only global attention & 25.00 & 24.54 & 32.87 & -0.84 \\
        \hline
    \end{tabular}
    \caption{Separate effect of group and global attention reporting in d-BLEU. Here, rnd. denotes the model trained using randomly initialized parameters.}
    \label{tab:combinedattn}
\end{table}

\subsection{Discussion of G-Transformer}
\textbf{Document Context.}
We study the contribution of the source-side and target-side context by removing the cross-sentential attention in Eq \ref{eq:combineattn} from the encoder and the decoder gradually.
The results are shown in Table \ref{tab:context}. We take the G-Transformer fine-tuned on the sentence-level Transformer as our starting point. When we disable the target-side context, the performance decreases by 0.14 s-BLEU point on average, which indicates that the target-side context does impact translation performance significantly. When we further remove the source-side context, the performance decrease by 0.49, 0.83, and 0.77 s-BLEU point on TED, News, and Europarl, respectively, which indicates that the source-side context is relatively more important for document-level MT.

To further understand the impact of the source-side context, we conduct an experiment on automatic evaluation on discourse phenomena which rely on source context. We use the human labeled evaluation set \cite{voita2019good} on English-Russion (En-Ru) for deixis and ellipsis. We follow the Transformer concat baseline \cite{voita2019good} and use both 6M sentence pairs and 1.5M document pairs from OpenSubtitles2018 \cite{lison2018opensubtitles} to train our model.
The results are shown in Table \ref{tab:discourse}. G-Transformer outperforms Transformer baseline concat \cite{voita2019good} with a large margin on three discourse features, indicating a better leverage of the source-side context. When compared to previous model LSTM-T, G-Transformer achieves a better ellipsis on both infl. and VP. However, the score on deixis is still lower, which indicates a potential direction that we can investigate in further study.

\textbf{Word-dropout.}
As shown in Table \ref{tab:ablation}, word-dropout (Appendix \ref{apdx:gtrans-training}) contributes about 0.37 d-BLEU on average. Its contribution to TED and News is obvious in 0.35 and 0.58 d-BLEU, respectively. However, for large dataset Europarl, the contribution drops to 0.17, suggesting that with sufficient data, word-dropout may not be necessary.

\textbf{Locality Bias.}
In G-Transformer, we introduce locality bias to the language modeling of source and target, and locality bias to the translation between source and target. We try to understand these biases by removing them from G-Transformer. When all the biases removed, the model downgrades to a document-level Transformer. The results are shown in Table \ref{tab:ablation}. Relatively speaking, the contribution of language locality bias is about 1.78 d-BLEU on average. While the translation locality bias contributes for about 14.68 d-BLEU on average, showing critical impact on the model convergence on small datasets. These results suggest that the locality bias may be the key to train whole-document MT models, especially when the data is insufficient.

\textbf{Combined Attention.}
In G-Transformer, we enable only the top K layers with combined attention. On Europarl7, G-Transformer gives 33.75, 33.87, and 33.84 d-BLEU with top 1, 2, and 3 layers with combined attention, respectively, showing that $K=2$ is sufficient.
Furthermore, we study the effect of group and global attention separately. As shown in Table \ref{tab:combinedattn}, when we replace the combined attention on top 2 layers with group attention, the performance drops by 0.22, 0.09, and 0.75 d-BLEU on TED, News, and Europarl, respectively. When we replace the combined attention with global attention, the performance decrease is enlarged to 0.84, 0.69, and 1.00 d-BLEU, respectively. These results demonstrate the necessity of combined attention for integrating local and global context information.

\section{Related Work}
The unit of translation has evolved from word \cite{brown1993mathematics,vogel1996hmm} to phrase \cite{koehn2003statistical,chiang2005hierarchical,chiang2007hierarchical} and further to sentence \cite{kalchbrenner2013recurrent,sutskever2014sequence,bahdanau2014neural} in the MT literature. The trend shows that larger units of translation, when represented properly, can lead to improved translation quality.

A line of document-level MT extends translation unit to multiple sentences  \cite{tiedemann-2017-extctx,agrawal2018-contextual,zhang2020long,Ma2020-flat}. However, these approaches are limited within a short context of maximum four sentences. Recent studies extend the translation unit to whole document
\cite{junczys-2019-msdoc,Liu2020-mBART}, using large augmented dataset or pretrained models.
\citet{Liu2020-mBART} shows that Transformer trained directly on document-level dataset can fail, resulting in unreasonably low BLEU scores.
Following these studies, we also model translation on the whole document.
We solve the training challenge using a novel locality bias with group tags.

Another line of work make document-level machine translation sentence by sentence, using additional components to represent the context \cite{maruf2018document,Zheng2020-towards,zhang-2018-improving,Miculicich2018-HAN,Maruf2019-SAN,yang2019enhancing}.
Different from these approaches, G-Transformer uses a generic design for both source and context, translating whole document in one beam search instead of sentence-by-sentence.
Some methods use a two-pass strategy, generating sentence translation first, integrating context information through a post-editing model \cite{voita2019-context,yu2020-better}. 
In contrast, G-Transformer uses a single model, which reduces the complexity for both training and inference.

The locality bias we introduce to G-Transformer is different from the ones in Longformer \cite{Beltagy2020Longformer} and Reformer \cite{Kitaev2020Reformer} in the sense that we discuss locality in the context of representing the alignment between source sentences and target sentences in document-level MT. Specifically, Longformer introduces locality only to self-attention, while G-Transformer also introduces locality to cross-attention, which is shown to be the key for the success of G-Transformer.
Reformer, basically same as Transformer, searches for attention targets in the whole sequence, while G-Transformer mainly restricts the attention inside a local sentence. In addition, the motivations are different. While Longformer and Reformer focus on the time and memory complexities, we focus on attention patterns in cases where a translation model fails to converge during training.

\section{Conclusion}
We investigated the main reasons for Transformer training failure in document-level MT, finding that target-to-source attention is a key factor. According to the observation, we designed a simple extension of the standard Transformer architecture, using group tags for attention guiding. Experiments show that the resulting G-Transformer converges fast and stably on small and large data, giving the state-of-the-art results compared to existing models under both pre-training and random initialization settings.

\section*{Acknowledgments}

We would like to thank the anonymous reviewers for their valuable feedback. We thank Westlake University High-Performance Computing Center for supporting on GPU resources. This work is supported by grants from Alibaba Group Inc. and Sichuan Lan-bridge Information Technology Co.,Ltd.

\bibliographystyle{acl_natbib}
\bibliography{acl2021}

\clearpage
\appendix

\section{Evaluation Metrics}
\label{apdx:metrics}
Following \citet{Liu2020-mBART}, we use sentence-level BLEU score (s-BLEU) as the major metric for our evaluation. However, when document-level Transformer is compared, we use document-level BLEU score (d-BLEU) since the sentence-to-sentence alignment is not available.

\textbf{s-BLEU.}
To calculate sentence-level BLEU score on document translations, we first split the translations into sentences, mapping to the corresponding source sentences. Then we calculate the BLEU score on pairs of translation and reference of the same source sentence.

\textbf{d-BLEU.} 
When the alignments between translation and source sentences are not available, we calculate the BLEU score on document-level, matching n-grams in the whole document.

\section{Transformer}
\label{apdx:transformer}

\subsection{Model}
\label{apdx:trans-model}

Transformer \cite{vaswani2017attention} has an encoder-decoder structure, using multi-head attention and feed-forward network as basic modules. In this paper, we mainly concern about the attention module.

\textbf{Attention.}
An attention module works as a function, mapping a query and a set of key-value pairs to an output, that the query, keys, values, and output are all vectors. The output is computed as a weighted sum of the values, where the weight assigned to each value is computed by a matching function of the query with the corresponding key.
Formally, for matrix inputs of query $Q$, key $K$, and value $V$,
\begin{equation}
\small
    \text{Attention}(Q, K, V) = \text{softmax} \left( \frac{QK^T}{\sqrt{d_k}} \right) V,
\end{equation}
where $d_k$ is the dimensions of the key vector.

\textbf{Multi-Head Attention.}
Build upon single-head attention module, multi-head attention allows the model to attend to different positions of a sequence, gathering information from different representation subspaces by heads.
\begin{equation}
\small
    \text{MultiHead}(Q, K, V) = \text{Concat}(head_1,...,head_h)W^O, \\
\end{equation}
where
\begin{equation}
\small
    head_i = \text{Attention}(Q W_i^Q, K W_i^K, V W_i^V),  
\end{equation}
that the projections of $W^O$, $W_i^Q$, $W_i^K$, and $W_i^V$ are parameter matrices.

\textbf{Encoder.}
The encoder consists of a stack of $N$ identical layers. Each layer has a multi-head self-attention, stacked with a feed-forward network. A residual connection is applied to each of them.

\textbf{Decoder.}
Similar as the encoder, the decoder also consists of a stack of $N$ identical layers. For each layer, a multi-head self-attention is used to represent the target itself, and a multi-head cross-attention is used to attend to the encoder outputs. The same structure of feed-forward network and residual connection as the encoder is used.

\subsection{Training Settings}
\label{apdx:trans-training}

We build our experiments based on Transformer implemented by Fairseq \cite{ott2019fairseq}. We use shared dictionary between source and target, and use a shared embedding table between the encoder and the decoder. 
We use the default setting proposed by Transformer \cite{vaswani2017attention}, which uses Adam optimizer with $\beta_1=0.9$ and $\beta_2=0.98$, a learning rate of $5e-4$, and an inverse-square schedule with warmup steps of $4000$.
We apply label-smoothing of 0.1 and dropout of 0.3 on all settings.
To study the impact of input length, data scale, and model size, we take the learning rate and other settings as controlled variables that are fixed for all experiments. We determine the number of updates/steps automatically by early stop on validation set.
We train base and big models on 4 GPUs of Navidia 2080ti, and large model on 4 GPUs of v100.

\begin{table*}[t]
    \centering\small
    \begin{tabular}{l|cc|cc|cc}
        \hline
        \bf{Method} & \multicolumn{2}{c|}{\bf{TED}} & \multicolumn{2}{c|}{\bf{News}} & \multicolumn{2}{c}{\bf{Europarl}} \\
         & s-BLEU & d-BLEU & s-BLEU & d-BLEU & s-BLEU & d-BLEU \\
        \hline
        G-Transformer random initialized (Base) & 23.53 & 25.84 & 23.55 & 25.23 & 32.18 & 33.87  \\
        G-Transformer random initialized (Big) & 23.29 & 25.48 & 22.22 & 23.82 & 32.04 & 33.77  \\
        G-Transformer random initialized (Large) & 6.23 & 8.95 & 13.68 & 15.33 & 31.51 & 33.21  \\
        \hline
    \end{tabular}
    \caption{G-Transformer on different model size.}
    \label{apdx:tab:compare-modelsize}
\end{table*}

\section{G-Transformer}
\subsection{Training Settings}
\label{apdx:gtrans-training}
We generate the corresponding group tag sequence dynamically in the model according to the special sentence-mark tokens {\bos} and {\eos}. Taking a document ``{\bos} there is no public transport . {\eos} {\bos} local people struggle to commute . {\eos}'' as an example, a group-tag sequence $G=\{1, 1, 1, 1, 1, 1, 1, 1, 2, 2, 2, 2, 2, 2, 2, 2\}$ is generated according to Eq \ref{eq:gxgy}, where $1$ starts on the first {\bos}  and ends on the first {\eos}, $2$ the second, and so on. The model can be trained either randomly initialized or fine-tuned.

\textbf{Randomly Initialized.}
We use the same settings as Transformer to train G-Transformer, using label-smoothing of $0.1$, dropout of $0.3$, Adam optimizer, and a learning rate of $5e-4$ with $4000$ warmup steps. To encourage inferencing the translation from the context, we apply a word-dropout \cite{bowman2016generating} with a probability of $0.3$ on both the source and the target inputs. 

\textbf{Fine-tuned on Sentence-Level Transformer.}
We use the parameters of an existing sentence-level Transformer to initialize G-Transformer. We copy the parameters of the multi-head attention in Transformer to the group multi-head attention in G-Transformer, leaving the global multi-head attention and the gates randomly initialized. For the global multi-head attention and the gates, we use a learning rate of $5e-4$, while for other components, we use a smaller learning rate of $1e-4$. All the parameters are jointly trained using Adam optimizer with $4000$ warmup steps. We apply a word-dropout with a probability of $0.1$ on both the source and the target inputs.

\textbf{Fine-tuned on mBART25.}
Similar as the fine-tuning on sentence-level Transformer, we also copy parameters from mBART25 \cite{Liu2020-mBART} to G-Transformer, leaving the global multi-head attention and the gates randomly initialized. We following the settings \cite{Liu2020-mBART} to train the model, using Adam optimizer with a learning rate of $3e-5$ and $2500$ warmup steps. Here, we do not apply word-dropout, which empirically shows a damage to the performance.

\subsection{Results on Model Size}
\label{apdx:gtrans-modelsize}

As shown in Table \ref{apdx:tab:compare-modelsize}, G-Transformer has a relatively stable performance on different model size. When increasing the model size from Base to Big, the performance drops for about 0.24, 1.33, and 0.14 s-BLEU points, respectively. Further to Large model, the performance drops further for about 17.06, 8.54, and 0.53 s-BLEU points, respectively. Although the performance drop on small dataset is large since overfitting on larger model, the drop on large dataset Europarl is relatively small, indicating a stable training on different model size.

\end{document}